\documentclass[10pt,twocolumn,letterpaper]{article}

\usepackage[pagenumbers]{cvpr} %

\usepackage[accsupp]{axessibility}
\usepackage[binary-units]{siunitx}
\usepackage{float}
\usepackage[utf8]{inputenc} %
\usepackage[T1]{fontenc}    %
\usepackage{url}            %
\usepackage{booktabs}       %
\usepackage{amsfonts}       %
\usepackage{nicefrac}       %
\usepackage{microtype}      %
\usepackage{mathtools}
\usepackage{graphicx}
\usepackage{amsmath}
\usepackage{amssymb}
\usepackage{amsthm}
\usepackage{colortbl}
\usepackage{wrapfig}
\usepackage{bm}
\usepackage[dvipsnames]{xcolor}
\usepackage{array}
\usepackage{enumitem}
\usepackage{multirow}
\usepackage{makecell}
\usepackage{siunitx}
\usepackage{algorithmic}
\usepackage{algorithm}
\usepackage{calc}
\makeatletter
\@namedef{ver@everyshi.sty}{}
\makeatother
\usepackage{tikz}
\usepackage{pgfplots}
\usepgfplotslibrary{units}
\usepackage{xspace}
\usepackage[colorlinks,breaklinks=true,pagebackref]{hyperref}
\hypersetup{linkcolor=BrickRed,citecolor=Green}
\pgfplotsset{compat=1.9}

\definecolor{Gray}{gray}{0.9}
\renewcommand{\algorithmiccomment}[1]{\textcolor{gray}{\bgroup\hfill//~#1\egroup}}

\usepackage{amsmath,amsfonts,bm}

\def\1{\bm{1}}
\newcommand{\train}{\mathcal{D}_{\text{train}}}
\newcommand{\valid}{\mathcal{D_{\mathrm{val}}}}
\newcommand{\test}{\mathcal{D_{\mathrm{test}}}}

\def\vlambda{{\bm{\lambda}}}
\def\vrho{{\bm{\rho}}}

\def\vl{{\bm{l}}}

\def\vs{{\bm{s}}}

\def\vx{{\bm{x}}}

\DeclareMathAlphabet{\mathsfit}{\encodingdefault}{\sfdefault}{m}{sl}
\SetMathAlphabet{\mathsfit}{bold}{\encodingdefault}{\sfdefault}{bx}{n}

\def\gD{{\mathcal{D}}}

\def\gX{{\mathcal{X}}}
\def\gY{{\mathcal{Y}}}

\def\sN{{\mathbb{N}}}

\def\sP{{\mathbb{P}}}

\newcommand{\Ls}{\mathcal{L}}
\newcommand{\R}{{\mathbb{R}}}

\newcommand{\softmax}{\mathrm{softmax}}

\newcommand{\N}{\mathbb{N}}

\DeclareMathOperator*{\argmax}{arg\,max}

\DeclareMathOperator*{\st}{s.t.}

\newcommand{\comment}[1]{}

\def\cvprPaperID{6055} %
\def\confName{CVPR}
\def\confYear{2023}

\begin{document}

\title{Class Adaptive Network Calibration}

\author{
Bingyuan~Liu\thanks{Equal Contributions. Correspondence to: \{\url{liubingyuan1988@gmail.com}, \url{jerome.rony.1@etsmtl.net}\}} $^{1}$,
Jérôme Rony$^{\ast1}$, Adrian~Galdran$^{2}$, Jose~Dolz$^{1}$, Ismail~Ben~Ayed$^{1}$\\[3mm]
$^{1}$ÉTS Montreal, Canada\hspace{0.5cm}  $^{2}$Universitat Pompeu Fabra, Barcelona, Spain %
}

\maketitle

\begin{abstract}
Recent studies have revealed that, beyond conventional accuracy, calibration should also be considered for training modern deep neural networks.
To address miscalibration during learning, some methods have explored different penalty functions as part of the learning objective, alongside a standard classification loss, with a hyper-parameter controlling the relative contribution of each term.
Nevertheless, these methods share two major drawbacks: 1) the scalar balancing weight is the same for all classes, hindering the ability to address different intrinsic difficulties or imbalance among classes; and 2) the balancing weight is usually fixed without an adaptive strategy, which may prevent from reaching the best compromise between accuracy and calibration, and requires hyper-parameter search for each application.
We propose Class Adaptive Label Smoothing (CALS) for calibrating deep networks, which allows to learn class-wise multipliers during training, yielding a powerful alternative to common label smoothing penalties.
Our method builds on a general Augmented Lagrangian approach, a well-established technique in constrained optimization, but we introduce several modifications to tailor it for large-scale, class-adaptive training. 
Comprehensive evaluation and multiple comparisons on a variety of benchmarks, including standard and long-tailed image classification, semantic segmentation, and text classification, demonstrate the superiority of the proposed method.
The code is available at \url{https://github.com/by-liu/CALS} .
\end{abstract}

\begin{figure*}[!ht]
\begin{center}
\includegraphics[width=.95\linewidth]{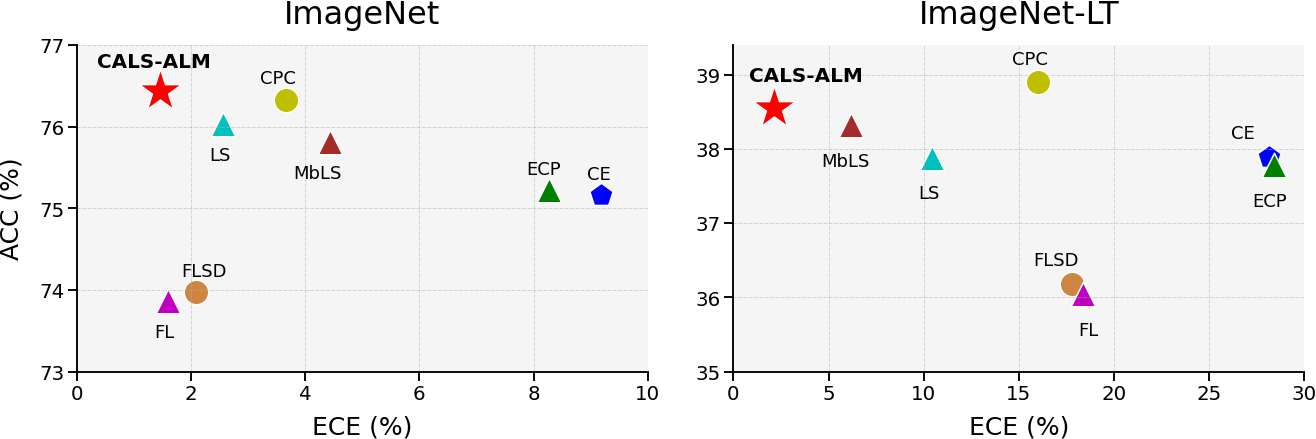}
\end{center}
\vspace{-1em}
\caption{Many techniques have been proposed for jointly improving accuracy and calibration during training\cite{guo2017calibration, mukhoti2020calibrating}, but they fail to consider uneven learning scenarios like high class imbalance or long-tail distributions. 
We show a comparison of the proposed \textbf{CALS-ALM} method and different learning approaches in terms of Calibration Error (ECE) vs Accuracy on the (a) ImageNet and (b) ImageNet-LT (long-tailed ImageNet) datasets.
A lower ECE indicates better calibration: a better model should attain \textbf{high ACC and low ECE}. 
Among all the considered methods, \textbf{CALS-ALM} shows superior performance when considering both discriminative power and well-balanced probabilistic predictions, achieving best accuracy and calibration on ImageNet, and best calibration and second best accuracy on ImageNet-LT.}
     \label{fig:overview}
\end{figure*}

\section{Introduction}
\label{sec:intro}

Deep Neural Networks (DNNs) have become the prevailing model in machine learning, particularly for computer vision\cite{he2016deep} and natural language processing applications\cite{Ashish2017Attention}.
Increasingly powerful architectures\cite{he2016deep, chen2017rethinking, liu2021Swin}, learning methods\cite{chen20simcl, he2022Masked} and a large body of other techniques   \cite{ioffe15batchnorm, loshchilov2018adamw} are constantly introduced.
Nonetheless, recent studies\cite{guo2017calibration, mukhoti2020calibrating} have shown that regardless of their superior discriminative performance, high-capacity modern DNNs are poorly calibrated, \ie failing to produce reliable predictive confidences.
Specifically, they tend to yield over-confident predictions, where the probability associated with the predicted class overestimates the actual likelihood.
Since this is a critical issue in safety-sensitive applications like autonomous driving or computational medical diagnosis, the problem of DNN calibration has been attracting increasing attention in recent years \cite{guo2017calibration,mukhoti2020calibrating,pereyra2017regularizing}.

Current calibration methods can be categorized into two main families.
The first family involves techniques that perform an additional post-processing parameterized operation on the output logits (or pre-softmax activations) \cite{guo2017calibration}, with the calibration parameters of that operation obtained from a validation set by either learning or grid-search.
Despite the simplicity and low computational cost, these methods have empirically proven to be highly effective \cite{guo2017calibration, Ding2021LocalTemp}.
However, their main drawback is that the choice of the optimal calibration parameters is highly sensitive to the trained model instance and validation set \cite{mukhoti2020calibrating, liu2022mbls}.

The second family of methods attempts to simultaneously optimize for accuracy and calibration during network training.
This is achieved by introducing, explicitly or implicitly, a secondary optimization goal involving the model's predictive uncertainty, alongside the main training objective.
As a result, a scalar balancing hyper-parameter is required to tune the relative contribution of each term in the overall loss function.
Some examples of this type of approaches include: Explicit Confidence Penalty (ECP)\cite{pereyra2017regularizing}, Label Smoothing (LS)\cite{muller2019does}, Focal Loss (FL) \cite{lin2017focal} and its variant, Sample-Dependent Focal Loss (FLSD) \cite{mukhoti2020calibrating}.
It has been recently demonstrated in\cite{liu2022mbls} that all these methods can be formulated as different penalty terms that enforce the same equality constraint on the logits of the DNN: driving the \textit{logit distances} towards zero.
Here, logit distances refers to the vector of L1 distances between the highest logit value and the rest.
Observing the non-informative nature of this equality constraint, \cite{liu2022mbls} proposed to use a generalized inequality constraint, only penalizing those logits for which the distance is larger than a pre-defined margin, achieving state-of-the-art calibration performance on many different benchmarks. %

Although learning based methods achieve greater calibration performance \cite{mukhoti2020calibrating, liu2022mbls}, they have two major limitations: 
1) The scalar balancing weight is equal for all classes. 
This hinders the network performance when some classes are harder to learn or less represented than others, such as in datasets with a large number of categories (ImageNet) or considerable class imbalance (ImageNet-LT). 
2) The balancing weight is usually fixed before network optimization, with no learning or adaptive strategy throughout training. 
This can prevent the model from reaching the best compromise between accuracy and calibration.
To address the above issues, we introduce Class Adaptive Label Smoothing method based on an Augmented Lagrangian Multiplier algorithm, which we refer to as CALS-ALM.
Our \textbf{Contributions} can be summarized as follows:
\begin{itemize}[leftmargin=*]
\item We propose Class Adaptive Label Smoothing (CALS) for network calibration.
Adaptive class-wise multipliers are introduced instead of the widely used single balancing weight, which addresses the above two issues: 1) CALS can handle a high number of classes with different intrinsic difficulties, \eg ImageNet; 2) CALS can effectively learn from data suffering from class imbalance or a long-tailed distribution, \eg ImageNet-LT.
\item Different from previous penalty based methods, we solve the resulting constrained optimization problem by implementing a modified Augmented Lagrangian Multiplier (ALM) algorithm, which yields adaptive and optimal weights for the constraints.
We make some critical design decisions in order to adapt ALM to the nature of modern learning techniques: 
1) The inner convergence criterion in ALM is relaxed to a fixed number of iterations in each inner stage, which is amenable to mini-batch stochastic gradient optimization in deep learning. 
2) Popular techniques, such as data augmentation, batch normalization \cite{ioffe15batchnorm} and dropout \cite{gal2016dropout}, rule out the possibility of tracking original samples and applying sample-wise multipliers. 
To overcome this complication, we introduce class-wise multipliers, instead of sample-wise multipliers in the standard ALM. 
3) The outer-step update for estimating optimal ALM multipliers is performed on the validation set, which is meaningful for training on large-scale training set and avoids potential overfitting. 
\item Comprehensive experiments over a variety of applications and benchmarks, including standard image classification (Tiny-ImageNet and ImageNet), long-tailed image classification (ImageNet-LT), semantic segmentation (PASCAL VOC 2012), and text classification (20 Newsgroups), demonstrate the effectiveness of our CALS-ALM method. 
As shown in \autoref{fig:overview}, CALS-ALM yields superior performance over baselines and state-of-the-art calibration losses when considering both accuracy and calibration, especially for more realistic large-scaled datasets with large number of classes or class imbalance.
\end{itemize}

\section{Related Work}
\label{sec:related_works}

\subsection{Problem Formulation}
\label{sec:problem_formulation}
Given a dataset $\gD=\{(\vx^{(i)}, y^{(i)})\}_{i=1}^N$ with ${N\in\sN}$ pairs of samples $\vx\in\gX$ and corresponding labels ${y\in\gY}$, with $\gY = \{1, \dots, K\}$, a deep neural network (DNN) $F_\theta:\gX \rightarrow \R^K$ parameterized by $\theta$ yields \emph{logits} ${\vl = F_\theta(\vx) = (F_\theta(\vx)_k)_{1\leq k\leq K}\in\R^K}$.
In a classification scenario, the output probability ${\vs = (s_k)_{1 \leq k \leq K} \in \Delta^{K-1}}$, where $\Delta^{K-1}\subset [0, 1]^K$ denotes the probability simplex, is obtained by applying the $\softmax$ function on the output \emph{logits}, \ie  ${\vs = \softmax(\vl) = \frac{\exp\vl}{\sum\exp{\vl}}}$.
Therefore, the predicted class $\hat{y}$ is computed as ${\hat{y} = \argmax_k s_k}$, and the predicted confidence is $\hat{p} = \vs_{\hat{y}} = \max_{k} s_k$.
A perfectly calibrated model should satisfy that the predicted confidence of any input is equal to the accuracy of the model: $\hat{p} = \sP(\hat{y} = y | \hat{p})$.
Hence, an over-confident model yields on average larger confidences than the associated accuracy, whereas an under-confident model yields lower confidence than its accuracy.

A number of recent studies \cite{guo2017calibration, mukhoti2020calibrating,  minderer2021revisiting, liu2022mbls} have shown that DNNs tend to become over-confident during training as a result of minimizing the popular cross-entropy (CE) training loss:
\begin{equation}
    \label{eq:ce}
    \Ls_\text{CE}(\vx, y) = - \sum_{k=1}^K y_k \log s_k
\end{equation}
where $\bm{y}\in\{0, 1\}^K$ is the one-hot encoding of $y$. 
This objective function is minimized when the predictions for all the training samples fully match the ground-truth labels $y$, \ie $s_y = 1$ and $\forall k\neq y, s_k = 0$.
The negative logarithmic term on the logit of the correct category renders the global minimization of the CE loss unreachable, as it keeps pushing the predicted probabilities $\vs$ towards the vertices of the $(K{-}1)$-simplex even after the classification error is zero \cite{mukhoti2020calibrating}, resulting in over-confident models.

\subsection{Post-processing methods}
To address mis-calibration, different post-processing techniques applied after model training have been proposed recently \cite{guo2017calibration,Ding2021LocalTemp,Tomani2021Posthoc}. 
The most popular of these strategies is Temperature Scaling (TS) \cite{guo2017calibration}, which applies a single scalar temperature parameter to manipulate logit outputs monotonely, resulting in softened prediction confidences without affecting predicted labels.
Note that here the temperature parameter needs to be tuned on a separate validation set.
Despite its simplicity, TS has been shown effective in fixing over-confidence predictions \cite{guo2017calibration}.
As a local alternative to TS, \cite{Ding2021LocalTemp} proposed to train a regression model for learning position-specific temperature for semantic segmentation problems.
Unfortunately, TS and its variants can be sensitive to both the model and the validation set, and do not work well under data distribution shifts \cite{ovadia2019can}.
Thus, some subsequent works \cite{Tomani2021Posthoc, Ma2021postrank} have attempted to provide solutions for improving performance under domain shift.

\subsection{Learning-based methods}
Another popular direction is to directly address mis-calibration during training by introducing an additional penalty or supervision regarding model calibration with the standard training loss.
In \cite{kumar2018trainable}, the authors introduced a trainable calibration measure based on RKHS kernels, while \cite{karandikar2021soft} proposed a differential calibration loss based on a soft version of the binning operation in the ECE metric.
In \cite{Cheng22CPC}, two types of binary pairwise calibration constraints were proposed as additional penalty terms during training.
Other methods try to decrease over-fitting on the cross-entropy loss, which has been demonstrated to be the main reason of over-confidence \cite{guo2017calibration,mukhoti2020calibrating}. 
In \cite{pereyra2017regularizing} an explicit confidence penalty (ECP) is proposed to maximize the entropy and reduce over-fitting, while Label Smoothing \cite{szegedy2016rethinking} has also been shown to implicitly improve the calibration \cite{muller2019does} by softening the hard one-hot targets in the cross-entropy.
The Focal Loss \cite{lin2017focal}, originally proposed to tackle class imbalance, can also be effective for calibration \cite{mukhoti2020calibrating}, as it implicitly minimizes the Kullback-Leibler (KL) divergence between the uniform distribution and the network softmax probabilities, thereby increasing the entropy of the predictions.
As an extension the Sample-Dependent Focal Loss (FLSD) was also proposed in \cite{mukhoti2020calibrating} to further boost calibration performance.

\paragraph{Margin-based Label Smoothing (MbLS).}A unifying constrained-optimization formulation of loss functions promoting calibration has been recently presented in \cite{liu2022mbls}.
Specifically, the additional penalties integrated in these methods, including ECP \cite{pereyra2017regularizing}, LS \cite{muller2019does} and FL \cite{mukhoti2020calibrating}, can be viewed as different forms of approximations to the same constraint, \ie enforcing the logit distances to be zero.
Noticing that this constraint is non-informative (its solution being uniformly distributed probabilities), \cite{liu2022mbls} further proposed a generalized formulation by relaxing the constraint to allow the logit distances being lower than a strictly positive margin. 

The specific formulation of MbLS \cite{liu2022mbls} is as follows. Given a margin $m\in\R_{+}$, the constrained optimization problem for network training is:
\begin{equation}
\label{eq:mbls_constraint}
\begin{aligned}
    \min_\theta \quad & \sum_{i=1}^N \Ls_\text{CE}(\vx^{(i)}, y^{(i)})\\
    \st \quad & \max_k\{\vl^{(i)}_k\} - l^{(i)} \preceq m\1_K, \quad i=1,\dots,N,
\end{aligned}
\end{equation}
where $\vl^{(i)} = F_\theta(\vx^{(i)})$. 
The minimum can be approximated by penalty-based optimization methods, transforming the above formulation into an unconstrained problem by means of simple ReLU functions: %
\begin{equation}
\label{eq:mbls_penalty}
\begin{aligned}
    \min_\theta \quad &\sum_{i=1}^N\Ls_\text{CE}(\vx^{(i)}, y^{(i)}) \:+\\
    \lambda &\sum_{i=1}^N\sum_{j=1}^K\max\{0, \max_k\{l^{(i)}_k\} - l^{(i)}_j - m\},
\end{aligned}
\end{equation}
where $\lambda\in\R_+$ is a scalar weight balancing contributions of the CE loss and the corresponding penalty.

\section{Sample-wise Constrained DNN Optimization}
\label{sec:solving_mbls}
Although MbLS can significantly improve calibration, the associated constrained problem \eqref{eq:mbls_constraint} is not solved accurately.
It is approximated by an unconstrained problem with a single uniform penalty, regardless of the data sample or category. 
However, the samples and classes considered in a classification problem have different intrinsic learning difficulties. 
Therefore, an improved training scheme would involve considering distinct penalty weights $\lambda$ for each sample and class. 
This would result in having to chose $N\times K$ penalty weights $\Lambda\in\R_+^{N\times K}$, with the resulting optimization problem being:
\begin{equation}
\label{eq:mbls_NK_penalties}
\begin{aligned}
    \min_\theta \quad &\sum_{i=1}^N\Ls_\text{CE}(\vx^{(i)}, y^{(i)}) \:+\\
    &\sum_{i=1}^N\sum_{j=1}^K \Lambda_{ij}\max\{0, \max_k\{l^{(i)}_k\} - l^{(i)}_j - m\}.
\end{aligned}
\end{equation}
From an optimization perspective, supposing that optimal weights $\theta^\star$ exists for problem \eqref{eq:mbls_constraint}, there exists $\Lambda^\star\in\R_+^{N\times K}$ such that $(\theta^\star, \Lambda^\star)$ is a saddle point of the Lagrangian associated to \eqref{eq:mbls_constraint}. These $\Lambda^\star$ are the Lagrange multipliers of the problem. 
Therefore, using $\Lambda = \Lambda^\star$ would be the best choice to solve \eqref{eq:mbls_NK_penalties}.

In practice, using the Lagrange multipliers of problem \eqref{eq:mbls_constraint} as the weights for the penalties may not be computationally feasible, and it could even result in degraded performance. 
Indeed, in the context of machine learning, we optimize a model's weights $\theta$ to solve \eqref{eq:mbls_constraint} on a training set $\train$, and expect to generalize on a test set $\test$, which we do not have access to during training. 
Because of the bias-variance trade-off, solving \eqref{eq:mbls_constraint} optimally would likely result in overfitting, \ie we may solve problem \eqref{eq:mbls_constraint} accurately on the train set, but not generalize properly on the test set, resulting in poor calibration and classification performance overall. 
This suggests that it could be preferable to evaluate during training the quality of multipliers on a separate validation set $\valid$.
Additionally, several mechanisms for training DNNs are not compatible with a straightforward minimization. 
First, the use of batch normalization yields predictions that are not independent between samples in a minibatch. 
Second, the use of regularization techniques such as dropout may lead to virtually inaccurate predictions on certain training samples, impacting the correct estimation of multipliers. 
Third, data augmentation, which is standard in DNN training, would result in additional penalty weights for the augmented samples: they can be easier or harder to classify than the original ones. 

In addition to the above obstacles, applying a penalty weight per sample and per class (resulting in $N\times K$ weights) would not scale well for large datasets and dense predictive tasks, such as semantic segmentation, which is typically formulated as a per-pixel classification task. 
Assuming that images in the dataset have a size of $H\times W$, this would result in $N\times H\times W\times K$ penalty weights. 
This rapidly becomes a limiting factor for moderately sized segmentation datasets. 
For instance, Pascal VOC 2012 \cite{VOC2015} contains 21 classes and 1464 training images, amounting to $2.62{\times}10^8$ pixels, or $5.5{\times}10^9$ penalty weights, which, stored as \texttt{float32}, would use ${\sim}\SI{20}{\gibi\byte}$. 
For Cityscapes \cite{cordts2016cityscapes}, containing approximately 3000 training images of size $2048{\times}1024$ in 19 classes, this amounts to ${\sim}\SI{445}{\gibi\byte}$.

Following the above observations, we introduce a relaxation of sample-wise penalties, and propose to solve the following problem:
\begin{equation}
\label{eq:mbls_K_penalties}
\begin{aligned}
    \min_\theta \quad &\sum_{i=1}^N\Ls_\text{CE}(\vx^{(i)}, y^{(i)}) \:+\\
    &\sum_{i=1}^N\sum_{j=1}^K \lambda_{j}\max\{0, \max_k\{l^{(i)}_k\} - l^{(i)}_j - m\},
\end{aligned}
\end{equation}
where $(\lambda_j)_{1\leq j\leq K}\in\R_+^K$. 
Since penalties are now class-wise, we need $K$ penalty weights. 
This has the advantage to scale well to denser classification tasks such as segmentation. 
However, we still face a challenging optimization problem (\ref{eq:mbls_K_penalties}), since we still need to chose $K$ weights, which can be extremely complicated for large-scale datasets with many classes such as ImageNet, which contains 1000 classes. 
In the next section, we introduce a numerical technique to deal with this challenge.

\section{Class Adaptive Network Calibration}
The challenge of the previous formulation stems from correctly choosing the weights $\vlambda\in\R_+^K$, which can be cumbersome when $K$ is large. Therefore, we propose to use an Augmented Lagrangian Multiplier (ALM) method to adaptively learn the weights of the penalties.

\subsection{General ALM}

ALM methods combine penalties and primal-dual updates to solve a constrained problem.
They have well-established advantages and enjoy widespread popularity in the general context of optimization \cite{dimi96lag, NoceWrig06, sangalli2021constrained}.
Specifically, we have the following generic constrained optimization problem:
\begin{equation}
    \min_x \quad f(x) \quad \st \quad h_i(x)\leq 0, \quad i=1,\dots,n
\end{equation}
where $f:\R^d\rightarrow \R$ is the objective function and $h_i:\R^d\rightarrow\R, i=1,\dots,n$ are the constraint functions. We tackle it by approximately solving a sequence $j\in\N$ of unconstrained problems:
\begin{equation}
\label{eq:lagrangian_alm}
    \min_x \quad \Ls^{(j)}(x) = f(x) + \sum_{i=1}^n P(h_i(x), \rho_i^{(j)}, \lambda_i^{(j)})
\end{equation}
with $P:\R\times \R_{++}^n\times \R_{++}^n \rightarrow \R$ a penalty-Lagrangian function, and $\vrho^{(j)}=(\rho_i)_{1\leq i\leq n}\in\R_{++}^n$, $\vlambda^{(j)}=(\lambda^{(j)}_i)_{1\leq i\leq n}\in\R_{++}^n$ the penalty parameters and multipliers associated to $P$ at the $j\text{-th}$ iteration. 
This sequence of unconstrained problems is called \emph{outer} iterations, while the steps in the minimization of $\Ls^{(j)}$ are called \emph{inner} iterations.

\begin{figure}
    \centering
    \begin{tikzpicture}[
        declare function={
            func(\x,\rho,\lambda) = \lambda / \rho * ((\rho * \x >= -0.5) * (0.5 * (\rho * \x)^2 + \rho * \x) + (\rho * \x < -0.5) * (-ln(-2 * \rho * \x) / 4 - 3 / 8));
        }]
        \begin{axis}[
                font=\footnotesize,
                xlabel={$z$}, ylabel={$P(z,\rho,\lambda)$},
                domain=-5:3, xmin=-5, xmax=3, ymin=-2, ymax=4,
                grid=major, grid style=dashed,
                samples=100,
                width=3in, height=2in,
                legend pos=north west, legend cell align=left,
            ]
            \addplot[MidnightBlue, thick] {func(x,1,1)};
            \addlegendentry{$\rho=1\hphantom{0}, \lambda=1$};
            \addplot[BurntOrange, thick] {func(x,10,1)};
            \addlegendentry{$\rho=10, \lambda=1$};
            \addplot[ForestGreen, thick] {func(x,1,0.3)};
            \addlegendentry{$\rho=1\hphantom{0}, \lambda=0.3$};
        \end{axis}
    \end{tikzpicture}
    \caption{A penalty-Lagrangian function $P$ with varying values of $\rho$ and $\mu$. Higher values of $\rho$ bring $P$ closer to an ideal penalty. The multiplier $\lambda$ is the derivative of $P$ \wrt the constraint at $z=0$.}
    \label{fig:penalty_lagrangian}
\end{figure}
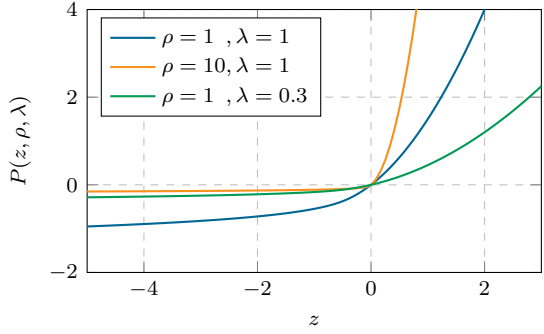

The main components of ALM methods are (\textit{i}) the \emph{penalty-Lagrangian function} $P$, (\textit{ii}) the update of the \emph{penalty multipliers} $\vlambda^{(j)}$ and (\textit{iii}) the increase of the \emph{penalty parameters} $\vrho^{(j)}$. First, the penalty function $P$ needs to satisfy a set of axioms \cite{birgin2005numerical} (see \autoref{sup:penalty_functions}): these axioms constrain the function to be continuously differentiable \wrt its first variable and to have a non-negative derivative: ${\forall z\in\R, P'(z,\rho,\lambda)=\frac{\partial}{\partial z}P(z,\rho,\lambda) \geq 0}$, with $P'(0, \rho, \lambda)=\lambda$. \autoref{fig:penalty_lagrangian} gives an example of a penalty, and how $\rho$ and $\lambda$ affect it. The choice of penalty function is critical to the performance of ALM methods, especially for nonconvex problems \cite{birgin2005numerical}.
Typical functions include PHR \cite{Hestenes1969MultiplierAG, Powell1969AMF}, $P_2$ \cite{Kort1976penalty} and $P_3$ \cite{Nakayama1975p3} (see section 3.2 of \cite{birgin2005numerical}).
Second, the penalty multipliers $\vlambda^{(j)}$ are updated to the derivative of $P$ \wrt the solution obtained during the last inner minimization.
Formally, let $x^{(j)}$ be the approximate minimizer of $\Ls^{(j)}$, then $\forall i\in\{1,\dots,n\}$:
\begin{equation}
\label{eq:multipliers_update_alm}
    \lambda_i^{(j+1)} = P'(h_i(x^{(j)}), \rho_i^{(j)}, \lambda_i^{(j)})
\end{equation}
This update rule corresponds to a first-order multiplier estimate for the constrained problem. Third, the penalty parameters $\vrho^{(j)}$ are increased during the outer iterations if the constraints do not improve (\ie is closer to being satisfied) compared to the previous outer iteration. Typically, $\rho_i^{(j+1)}=\gamma \rho_i^{(j)}$ if $h_i$ does not improve, with $\gamma>1$.

When the problem is convex, alternating between the approximate minimization of \eqref{eq:lagrangian_alm} and the update of the multipliers \eqref{eq:multipliers_update_alm} leads to a solution for the constrained problem.
The inner minimization corresponds to minimizing the primal while the outer iterations correspond to solving the dual problem.
The complete procedure is presented in \autoref{alg:general_alm}.
Although guarantees exist only in the convex case, it is well-known that ALM methods can efficiently solve nonconvex problems as well \cite{birgin2005numerical}. In the context of deep learning, their use has been surprisingly underexplored \cite{rony2020augmented,sangalli2021constrained}.

\begin{algorithm}
    \caption{Augmented Lagrangian Multiplier algorithm}
    \small
    \label{alg:general_alm}
    \begin{algorithmic}[1] 
        \REQUIRE Objective function $f$
        \REQUIRE Constraint functions $h_i, i=1,\dots,n$
        \REQUIRE Penalty function $P$, initial $\bm{\lambda}^{(0)}
        \in\R_{++}^n$,  $\bm{\rho}^{(0)}\in\R_{++}^n$
        \REQUIRE Initial variable $x^{(0)}$, iterations $j = 1$
        \WHILE{not converged}
            \STATE Initialize with $x^{(j-1)}$ and minimize (approximately):
            
            $\Ls^{(j)}(x) = f(x) + \sum_{i=1}^n P(h_i(x), \rho_i^{(j)}, \lambda_i^{(j)})$
            \STATE $x^{(j)} \gets$ (approximate) minimizer of $\Ls^{(j)}$
            \FOR{$i=1,\dots,n$}
                \STATE $\lambda_i^{(j+1)} \gets P'(h_i(x^{(j)}), \rho_i^{(j)}, \lambda_i^{(j)})$
                \IF{the $i$-th constraint does not improve} 
                    \STATE $\rho_i^{(j+1)}\gets \gamma \rho_i^{(j)}$
                \ELSE
                    \STATE $\rho_i^{(j+1)}\gets \rho_i^{(j)}$
                \ENDIF
            \ENDFOR
            \STATE $j \gets j + 1$
        \ENDWHILE
    \end{algorithmic}
\end{algorithm}

\subsection{ALM for calibration}
\label{sec:alm_calib}

Our goal now is to build an ALM method effective for calibration purposes. 
We can achieve this by reformulating problem \eqref{eq:mbls_K_penalties} using a penalty function $P$ parameterized by $(\vrho, \vlambda)\in\R_{++}^K\times\R_{++}^K$ as follows:
\begin{equation}
\label{eq:alm_mbls_unnormalized}
    \min_\theta \quad \sum_{i=1}^N\Ls_\text{CE}(\vx^{(i)}, y^{(i)})+\sum_{k=1}^K P(d_k^{(i)} - m, \rho_k, \lambda_k)
\end{equation}
where $d_k^{(i)} = \max\{\vl^{(i)}\} - l^{(i)}_k \in \R_+$. With this formulation, it is natural to use a penalty-Lagrangian function for $P$. To avoid numerical issues typically associated with non-linear penalties, we normalize the constraints by the margin $m>0$:
\begin{equation}
\label{eq:normalized_margin}
    d_k^{(i)} - m \leq 0 \Leftrightarrow \frac{d_k^{(i)}}{m} - 1 \leq 0
\end{equation}
This leads to improved numerical stability for the ALM multiplier update as well. Additionally, we average the constraints instead of summing them. This makes the method independent of the number of classes, and eases the choice of initial penalty parameters $\vrho^{(0)}$. The resulting loss is:
\begin{equation}
\label{eq:alm_mbls_normalized}
    \sum_{i=1}^N\Ls_\text{CE}(\vx^{(i)}, y^{(i)})+\frac{1}{K}\sum_{k=1}^K P\bigg(\frac{d_k^{(i)}}{m} - 1, \rho_k, \lambda_k\bigg)
\end{equation}

As noted in \autoref{sec:solving_mbls}, 
one of the main downsides of estimating Lagrange multipliers from the training set is that we could quickly overfit the data. Therefore, we propose to use the validation set to obtain a reliable estimate of the penalty multipliers at each epoch. We consider that an epoch of training corresponds to the approximate minimization of the loss function, and then compute the average penalty multiplier on the validation set. Formally after a training epoch $j$, the penalty multipliers for epoch $j{+}1$ will be, for all $k=1,\dots,K$:
\begin{equation}
    \lambda_k^{(j+1)} = \frac{1}{|\valid|}\smashoperator[r]{\sum_{(\vx,y)\in\valid}}P'\left(\frac{d_k}{m} - 1, \rho_k^{(j)}, \lambda_k^{(j)}\right)
\end{equation}
Finally, the penalty multiplier is projected on a safeguarding interval $[\lambda_{\min}, \lambda_{\max}]=[10^{-6}, 10^6]$ in our case. To update the penalty parameters $\vrho$, we compute the average constraint per class on the validation set. Then, for each class, if the average constraint is positive and has not decreased compared to the previous epoch, we multiply the corresponding penalty parameter by $\gamma$.

Finally, as suggested by \cite{birgin2005numerical} and confirmed by our empirical results, we utilize the PHR function in our implementation, defined as follows:
\begin{equation}
\label{eq:phr}
    \mathrm{PHR}(z,\rho,\lambda) =
    \begin{cases}
        \lambda z + \frac12 \rho z^2 \quad &\text{if} \quad \lambda + \rho z \geq 0;\\
        -\frac{\lambda^2}{2\rho} \quad &\text{otherwise}.
    \end{cases}
\end{equation}
Overall, the proposed method, consolidated in \autoref{alg:alm_training}, corresponds to approximately solving the constrained problem \eqref{eq:mbls_constraint}, by learning class-wise penalty multipliers on the validation set, to avoid overfitting and training specificities (\ie batch normalization, dropout, augmentations) which would result in unreliable penalty multipliers estimates.

\begin{algorithm}
    \caption{CALS-ALM training}
    \small
    \label{alg:alm_training}
    \begin{algorithmic}[1]
        \REQUIRE DNN initial $\theta^{0}$, margin $m$
        \REQUIRE Dataset: $\train$, $\valid$, batch size $B$
        \REQUIRE Penalty function $P$, $\gamma>1$, $\tau\in(0,1)$
        \REQUIRE Initial $\vlambda^{(0)}
        \in\R_{++}^n$,  $\vrho^{(0)}\in\R_{++}^n$
        \FOR[Epochs of training]{$j=0,\dots,T$}
            \FOR{each mini-batch $\{(\vx^{(i)}, y^{(i)})\}_{i=1}^B$ in $\train$}
                \STATE $\Ls_\text{c} = \sum\limits_{i=1}^B\Ls_\text{CE}(\vx^{(i)}, y^{(i)})$ \COMMENT{Cross-entropy}
                \STATE $\Ls_\text{p} = \sum\limits_{i=1}^B\frac{1}{K}\sum\limits_{k=1}^K P\Big(\frac{d_k^{(i)}}{m} - 1, \rho^{(j)}_k, \lambda^{(j)}_k\Big)$  \COMMENT{Penalties}
                \STATE $\Ls = \frac{1}{B}(\Ls_\text{c} + \Ls_\text{p})$
                \STATE $\theta^{(t+1)} \gets \theta^{(t)} - \alpha \cdot \nabla_\theta \Ls$  \COMMENT{Gradient descent}
            \ENDFOR
            \FOR[Adjust $\vlambda$ and $\vrho$ on validation]{$k=1,\dots,K$}
                \STATE $\lambda_k^{(j+1)} = \frac{1}{|\valid|}\smashoperator[r]{\sum_{(\vx,y)\in\valid}}P'\left(\frac{d_k}{m} - 1, \rho_k^{(j)}, \lambda_k^{(j)}\right)$
                \STATE $\overline{d_k}^{(j)} = \frac{1}{|\valid|}\smashoperator[r]{\sum_{(\vx,y)\in\valid}}\frac{d_k}{m} - 1$  \COMMENT{Average constraint}
                \IF{$j\geq 1$ \textbf{and} $\overline{d_k}^{(j)} > \tau \max\{0, \overline{d_k}^{(j-1)}\}$}
                    \STATE $\rho^{(j+1)}_k \gets \gamma \rho^{(j)}_k$  \COMMENT{Constraint has not improved}
                \ELSE
                    \STATE $\rho^{(j+1)}_k \gets \rho^{(j)}_k$
                \ENDIF
            \ENDFOR
        \ENDFOR
    \end{algorithmic}
\end{algorithm}

\section{Experiments}
\label{sec:exp}

\begin{table*}
    \centering
    \resizebox{0.9\textwidth}{!}{
        \begin{tabular}{l@{\hskip 3em} *{19}{c}}
            & \multicolumn{3}{c}{TinyImageNet} & & \multicolumn{7}{c}{ImageNet} & &  \multicolumn{7}{c}{ImageNet-LT}\\
            \cmidrule(l{1em}r{1em}){2-4}\cmidrule(l{1em}r{1em}){6-12}\cmidrule(l{1em}r){14-20}
            & \multicolumn{3}{c}{ResNet-50} & & \multicolumn{3}{c}{ResNet-50} & & \multicolumn{3}{c}{SwinV2-T} & &  \multicolumn{3}{c}{ResNet-50} & & \multicolumn{3}{c}{SwinV2-T}\\
            \cmidrule{2-4} \cmidrule{6-8} \cmidrule{10-12} \cmidrule{14-16} \cmidrule{18-20}
            Method & Acc & ECE & AECE & & Acc & ECE & AECE & & Acc & ECE & AECE & & Acc & ECE & AECE & & Acc & ECE & AECE \\
            \midrule
            CE                                      & 65.02 & 3.73 & 3.69 & & 75.16 & 9.19 & 9.18 & & 75.60 & 9.95 & 9.94 & & 37.90 & 28.12 & 28.12 & & 31.82 & 31.82 & 36.68 \\
            MMCE \cite{kumar2018trainable}          & \underline{65.34} & 2.81 & 2.61 & & 74.85 & 8.57 & 8.56 & & 76.68 & 9.07 & 9.08 & & 37.79 & 28.41 & 28.40 & & 33.14 & 26.41 & 26.41 \\
            ECP \cite{pereyra2017regularizing}      & 64.90 & 4.00 & 3.92 & & 75.22 & 8.27 & 8.26 & & 75.82 & 9.88 & 9.86 & & 37.69 & 28.14 & 28.13 & & 31.22 & 33.70 & 33.70 \\
            LS \cite{szegedy2016rethinking}         & \textbf{65.78} & 3.17 & 3.16 & & 76.04 & 2.57 & 2.88 & & 75.42 & 7.32 & 7.33 & & 37.88 & 10.46 & 10.38 & & 31.70 & 11.42 & 11.40 \\
            FL \cite{mukhoti2020calibrating}        & 63.09 & 2.96 & 3.12 & & 73.87 & \underline{1.60} & \underline{1.65} & & 75.60 & 3.19 & 3.18 & & 36.04 & 18.37 & 18.36 & & 30.73 & 25.50 & 25.50 \\
            FLSD \cite{mukhoti2020calibrating}      & 64.09 & 2.91 & 2.95 & & 73.97 & 2.08 & 2.06 & & 74.70 & 2.44 & 2.37 & & 36.18 & 17.77 & 17.78 & & 32.56 & 25.16 & 25.17  \\
            CPC \cite{Cheng22CPC}                   & 64.49 & 4.88 & 4.91 & & 76.33 & 3.66 & 3.59 & & 76.34 & 5.50 & 5.33 & & \textbf{38.90} & 16.00 & 15.99 & & 32.54 & 13.21 & 13.19 \\
            MbLS \cite{liu2022mbls}                 & 64.74 & \underline{1.64} & \underline{1.73} & & 75.82 & 4.44 & 4.26 & & \underline{77.18} & \underline{1.95} & \underline{1.73} & & 38.32 & 6.16 & 6.16 & & 32.05 & 7.65 & 7.64 \\
            \midrule
            CALS-HR                                 & 65.09 & 2.50 & 2.42 & & \underline{76.34} & 5.63 & 5.69 & & \textbf{77.58} & 3.06 & 2.95 & & 38.50 & \underline{2.83} & \underline{2.78} & & \textbf{34.31} &  \underline{2.37} & \textbf{2.45} \\
            \textbf{CALS-ALM}                       & 65.03 & \textbf{1.54} &  \textbf{1.38} & & \textbf{76.44} & \textbf{1.46} & \textbf{1.32} & & 77.10 & \textbf{1.61} & \textbf{1.69} & & \underline{38.56} & \textbf{2.15} & \textbf{2.30} & & \underline{33.94} & \textbf{2.32} & \textbf{2.45} \\
            \bottomrule
        \end{tabular}
    }
    \caption{Calibration performance for different approaches on three image classification benchmarks. We report two lower-is-better calibration metrics, \ie ECE and AECE. Best method is highlighted in bold, while the second-best one is underlined.}
    \label{tab:big}
    \vspace{-3mm}
\end{table*}

\subsection{Experimental Setup}

\noindent \textbf{Datasets.} We perform experiments on a variety of popular benchmarks. 
First, we include three widely used image classification benchmarks, including \textit{Tiny-ImageNet} \cite{deng2009imagenet}, \textit{ImageNet} \cite{deng2009imagenet} and one long-tailed image classification, \textit{ImageNet-LT} \cite{Liu2019imagenetlt}.
Tiny-ImageNet is widely used in the calibration literature \cite{mukhoti2020calibrating, liu2022mbls}, with relatively small $64{\times}64$ resolution, while ImageNet \cite{deng2009imagenet} is a large-scale benchmark consisting of $1000$ categories and over 1M images.
The main characteristic of ImageNet-LT is that the number of samples is extremely imbalanced across classes, ranging from $5$ to $1280$.
To evaluate performance in dense prediction tasks, we include one semantic segmentation benchmark, \textit{PASCAL VOC2012} \cite{VOC2015}.
Furthermore, one benchmark from the NLP domain, \textit{20 Newsgroups} \cite{lang1995newsweeder}, is included to show the general applicability.
For a detailed description of each dataset and the pre-processing settings, please refer to \autoref{sup:dataset}.

\noindent \textbf{Evaluation Metrics.} 
For calibration, we report the most widely used Expected Calibration Error (ECE)\cite{naeini2015ece}.
Samples are grouped into $M$ equi-spaced bins according to prediction confidence, and a weighted average of the absolute difference between accuracy and confidence in each bin is calculated:
\begin{align}\label{eq:ece}
ECE =  \sum_{m=1}^{M} \frac{|B_m|}{N} |A_m - C_m|,
\end{align}
where $M$ is the number of bins, $N$ the amount of test samples, $B_{m}$ the samples with prediction confidence in the $m^{th}$ bin, $A_m$ the accuracy and $C_m$ the mean confidence of samples in the $m^{th}$ bin.
Note we fix $N$ to $15$ according to \cite{mukhoti2020calibrating, liu2022mbls}.
In accordance with \cite{liu2022mbls}, we also report Adaptive ECE (AECE), a variant of ECE  where the bins are configured to evenly distribute the test samples across them.
Additionally, Classwise Calibration Error (CWCE) \cite{MaierHein2022MetricsRP}, a classwise extension of ECE, is included in \autoref{sup:add}.
For discriminative performance, we use standard measures: accuracy (Acc) for classification, and intersection over union (mIoU) for segmentation.

\noindent \textbf{Compared methods.} 
We compare our method to other learning based calibration losses, including \textit{(i)} methods that impose constraints on predictions (either logits or softmax probabilities), \ie Explicit Confidence Penalty (ECP) \cite{pereyra2017regularizing}, Label Smoothing (LS) \cite{szegedy2016rethinking}, Focal Loss (FL) \cite{lin2017focal} and its sample-dependent version (FLSD) \cite{mukhoti2020calibrating}, Margin-based Label Smoothing (MbLS) \cite{liu2022mbls} and CPC \cite{Cheng22CPC}, and \textit{(ii)} techniques that directly optimize calibration measures, \ie MMCE \cite{kumar2018trainable}.
We refer to the related literature \cite{mukhoti2020calibrating, liu2022mbls} to set the hyper-parameters for various methods.
For instance, the smoothing factor in LS and FL is set to $0.05$ and $3$ respectively, and we set margin to $10$ in MbLS. 
A detailed description of hyper-parameter values can be found in \autoref{sup:hyper}.

\noindent \textbf{Our methods.} 
A simple alternative to the algorithm presented in \autoref{sec:alm_calib} would be to heuristically tune multipliers by scaling them according to penalty values: if $P_k^{(j+1)}$ increases we also increase $\lambda_k^{j+1}$ and vice versa. 
This strategy, akin to learning rate scheduling, can be formulated as:
\begin{equation}\label{eq:heuristic}
    \lambda_k^{(j+1)} = 
    \begin{cases}
        \mu \lambda_k^{(j)}     & \text{if}\quad P_k^{(j+1)} > \tau  P_k^{(j)} \\[1.25mm]
        \lambda_k^{(j)}/\mu & \text{if}\quad P_k^{(j)} > \tau P_k^{(j+1)} \\[1.25mm]
        \lambda_k^{(j)} & \text{otherwise}
    \end{cases}
\end{equation}
where $\mu > 1$ and $\tau > 1$ are hyper-parameters that we fix to $1.1$.
We refer to our main algorithm as \textbf{CALS-ALM} and to this heuristic rule as \textbf{CALS-HR} in what follows. 

We fix the margin to $m=10$ on vision tasks and $m=6$ on the NLP benchmark, as in \cite{liu2022mbls}, for a fair comparison. 
We also perform an ablation study to investigate the impact of the margin value.
For other hyper-parameters, we set $\vlambda^{(0)}=10^{-6}\cdot\1_K$, $\vrho^{(0)}=\1_K$, $\gamma=1.2$, and we update the penalty parameters $\vrho$ every $10$ epochs. 
Please refer to \autoref{sup:hyper} for a detailed description of all hyper-parameters.

\noindent \textbf{Implementation Details.}
For image classification, we experiment with ResNet \cite{he2016deep} and a vision Transformer model, \ie Swin Transformer V2 (SwinV2-T) \cite{liu2021swinv2}.
DeepLabV3 \cite{chen2017rethinking} is employed for semantic segmentation on PASCAL VOC2012.
Following \cite{mukhoti2020calibrating, liu2022mbls}, we use the Global Pooling CNN (GPool-CNN) architecture \cite{lin2013network} on the NLP recognition task.
Further training details on each dataset can be found in \autoref{sup:dataset}.

\subsection{Results}

\begin{figure*}
    \centering
        \begin{subfigure}{2.375in}
        \includegraphics{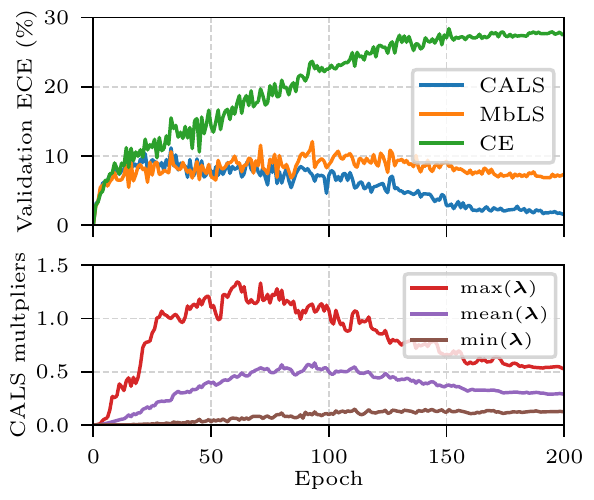}
        \caption{Validation ECE and multipliers during training.}
    \end{subfigure}
    \hfill
    \begin{subfigure}{4.25in}
        \includegraphics{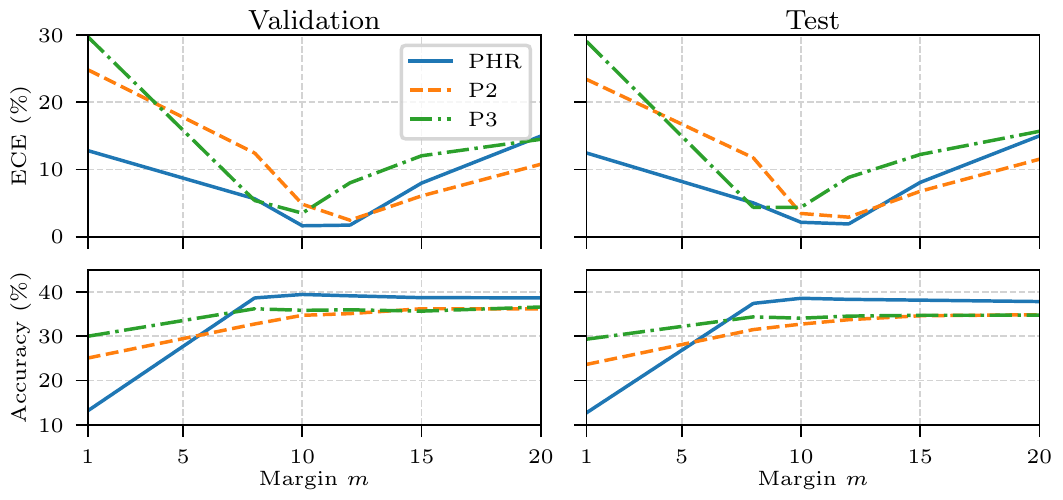}
        \caption{Effect of penalty function and margin on accuracy and ECE for CALS.}
    \end{subfigure}
    \caption{\textbf{Ablation study on ImageNet-LT.} (a) \textbf{Evolution of ECE on validation and multipliers for CALS}: 
    ECE on the validation set for our method (CALS), CE and MbLS \cite{liu2022mbls} and values of multipliers $\vlambda$ for CALS after each training epoch. (b) \textbf{Effect of penalty functions and margin}: ECE and accuracy on validation and test set are shown across different choices of penalty functions and margin values.}
    \label{fig:ablation}
\end{figure*}

\noindent \textbf{Results on image classification.} \autoref{tab:big} presents the discriminative and calibration performance of our methods across three widely used classification benchmarks, compared to baselines and related works. 
We can observe that our CALS-ALM approach consistently outperforms existing techniques in terms of calibration.
Specifically, the results indicate that the standard CE loss and other approaches often lead to miscalibrated models, with the severity of miscalibration substantially increasing in correlation with dataset difficulty.
This is particularly evident in large-scaled datasets with numerous classes, such as ImageNet, or those with long-tail class distributions, like ImageNet-LT.
Although other learning-based methods could provide better calibrated networks, their performance is not stable across different settings.
For example, while FL achieves a relatively low ECE of $1.60$ with a Resnet50 trained on ImageNet, it only yields an ECE of $25.50$ when training a SwinV2-T network on ImageNet-LT, revealing a limitation in adapting to different learning scenarios.
In contrast, CALS-ALM attains the best calibration performance in all cases, frequently outperforming exsiting approaches by a significant margin, with minimal variations across datasets and architectures. 
This trends persists when compared to the most closely related technique, MbLS.
It is noteworthy that the improvements on the long-tailed ImageNet-LT dataset are substantial; for example, we decrese ECE from 28.12 to 2.15 for ResNet-50, and from 31.82 to 2.32 for SwinV2-T, validating the effectiveness of class adaptive learning.
Another interesting finding is that CALS-HR, employing a naive update strategy for class-wise penalty weights, achieves nearly the second-best performance in the ImageNet-LT dataset.
This further demonstrates the effectiveness of CALS for learning under class-imbalance scenarios.
For the reliability diagrams of various models, please refer to \autoref{sup:reliability}.

In terms of model accuracy, our method delivers competitive performances, surpassing existing methods in certain cases. 
It is important to emphasize that, while the proposed method achieves discriminative results comparable to the best-performing approach for each dataset, the differences in calibration are considerable. 
This highlights the superiority of the proposed formulation for training highly discriminative and well-calibrated networks. \autoref{fig:overview} provides a more visual comparison considering both accuracy and ECE.
It is demonstrated that CALS-ALM provides the optimal compromise between accuracy and calibration performance.

\textbf{Ablation Analysis.} \autoref{fig:ablation} illustrates the evolution of ECE in (a) and penalty multipliers $\vlambda$ in (b) during training.
It is interesting to observe that the evolution of $\vlambda$ is consistent with the ECE. 
Specifically, the average penalty weight gradually increases while the ECE initially deteriorates because the model is focused on increasing accuracy. 
However, the value of the penalty weight begins to decline alongside the ECE, as the network starts to become better calibrated.
For a visualization of classwise multipliers, please refer to \autoref{sup:vis-class-lambd}.
\autoref{fig:ablation}(c) highlights the impact of the choice of penalty functions and margin values.
This demonstrates that the PHR penalty function is preferable over the other two options, P2 and P3, for both calibration and accuracy.
Regarding the margin, the best performance is achieved with $m\approx10$, which is consistent with the findings in \cite{liu2022mbls}.

\textbf{Semantic Segmentation.} \autoref{tab:voc} presents the performance on Pascal VOC dataset.
Note here CALS refers to our best method, \ie CALS-ALM.
It can be observed that the trend is consistent with image classification experiments: CALS outperforms counterparts in terms of ECE, and yields competitive results on discriminative performance, \ie mIoU in segmentation.
It is worth noting that some methods like MMCE, FLSD and CPC, are not included here because their computation demands were too heavy for pixel-wise segmentation tasks. In contrast, our method is unlimited in dense prediction tasks as the computation cost it adds is moderate.

\begin{table}
    \centering
    \small 
    \resizebox{0.8\columnwidth}{!}
    {
    \begin{tabular}{lccccccc}
        & CE & ECP & LS & FL & MbLS & CALS \\
        \midrule
        ECE & 14.75 & \underline{5.15} & 6.90 & 10.87 & 5.22 & \textbf{4.66} \\
        \midrule
        mIoU & 66.46 & 65.57 & \textbf{67.73} & 64.25 & 65.29 & \underline{66.77} \\
        \bottomrule
    \end{tabular}
    }
    \caption{Segmentation results on the PASCAL VOC 2012.}
    \label{tab:voc}
\end{table}

\textbf{Text Classification.} Last, we demonstrate the general applicability of the proposed method by analyzing its performance on a non-vision task, \ie text classification on 20 Newsgroups dataset.
The results are reported in \autoref{tab:20news}.
Remarkably, CALS again brings substantial improvement in terms of calibration, with ECE decreasing to $2.04\%$, while yielding the best accuracy $68.32\%$.
This reveals that the proposed class adaptive learning method is also able to handle class differences in NLP applications and provide promising performance in terms of both accuracy and calibration.

\begin{table}
    \centering
    \small
    \setlength{\tabcolsep}{4pt}
    \resizebox{\columnwidth}{!}{
    \begin{tabular}{lccccccccc}
        & CE & MMCE & ECP & LS & FL & FLSD & CPC & MbLS & CALS \\
        \midrule
        ECE (\%) & 22.75 & 23.02 & 22.97 & 8.07 & 10.80 & 10.87 & 9.46 & \underline{5.40} & \textbf{2.04} \\
        \midrule
        Acc (\%) & 67.01 & 66.23 & 66.48 & 67.14 & 66.08 & 65.85 & \underline{68.27} & 67.89 & \textbf{68.32} \\
        \bottomrule
    \end{tabular}
    }
    \caption{Results on the text classification task, 20 Newsgroups.}
    \label{tab:20news}
    \vspace{-3mm}
\end{table}

\section{Limitations and Future Work}
We have proposed Class Adaptive Label Smoothing for network calibration based on a modified Augmented Lagrangian Multiplier algorithm.
Despite its superior performance over previous methods, there are potential limitations in this work.
For instance, our method requires the validation set to have the same distribution as the training set.
Although this is satisfied in nearly every benchmark, it will be interesting to investigate the impact of using non-independent and identically distributed (\textit{i.i.d}) validation sets.

\section*{Acknowledgements}

This work is supported by National Science and Engineering Research Council of Canada (NSERC) and Prompt Quebec.
We also thank Calcul Quebec and Compute Canada.
Adrian Galdran is supported by a Marie Sklodowska-Curie Fellowship (No 892297).

\clearpage
{\small
\bibliographystyle{ieee_fullname}
\bibliography{egbib}
}

\clearpage
\appendix
\section{Penalty functions for ALM}
\label{sup:penalty_functions}

Here, we provide the requirements for a penalty function in Augmented Lagrangian Multiplier (ALM) method.

A function $P : \R \times \R_{++} \times \R_{++} \rightarrow \R $ is a Penalty-Lagrangian function such that $P'(z, \rho, \lambda) = \frac{\partial}{\partial y}P(z, \rho, \lambda)$ exists and is continuous for all $z \in \R$, $\rho \in \R_{++}$ and $\lambda \in \R_{++}$. 
In addition, it should satisfy the following four axioms \cite{birgin2005numerical}:
\begin{itemize}[leftmargin=*,label={}]
    \item {\bf Axiom 1:} $P'(z, \rho, \lambda) \geq 0 \quad \forall z\in\R,  \rho \in \R_{++}, \lambda \in \R_{++}$
    \item {\bf Axiom 2:} $P'(z, \rho, \lambda) = \lambda  \quad \forall \rho \in \R_{++}, \lambda \in \R_{++}$
    \item {\bf Axiom 3:} If, for all $j\in\N$, $0 < \lambda_\text{min} \leq \lambda^{(j)} \leq\lambda_\text{max} < \infty$, then:
    $\lim\limits_{j\rightarrow\infty}\rho^{(j)}=\infty$ and $\lim\limits_{j\rightarrow\infty}z^{(j)}>0$ imply that $\lim\limits_{j\rightarrow\infty}P'(z^{(j)}, \rho^{(j)}, \lambda^{(j)})=\infty$
    \item {\bf Axiom 4:} If, for all $j\in\N$, $0 < \lambda_\text{min} \leq \lambda^{(j)} \leq\lambda_\text{max} < \infty$, then: $\lim\limits_{j\rightarrow\infty}\rho^{(j)}=\infty$ and $\lim\limits_{j\rightarrow\infty}z^{(j)}<0$ imply that $\lim\limits_{j\rightarrow\infty}P'(z^{(j)}, \rho^{(j)}, \lambda^{(j)})=0$
\end{itemize} 
where the first two axioms guarantee the derivative of the Penalty-Lagrangian function $P$ \wrt $z$ is positive and equals to $\lambda$ when $z=0$, while the last two axioms guarantee that the derivative tends to infinity when the constraint is not satisfied, and zero when the constraint holds.

There are many valid penalty functions \cite{birgin2005numerical}.
In this paper we adopt the PHR function suggested by \cite{birgin2005numerical} and confirmed by our empirical results in Section 5 of the main text.
We also empirically compare with another two popular choices, \ie P2 and P3 \cite{birgin2005numerical}, as shown in Figure 3 of the main text.
The formulations of the above three penalty functions are as follows:
\begin{equation}
    \mathrm{PHR}(z,\rho,\lambda) =
    \begin{cases}
        \lambda z + \frac12 \rho z^2 \quad &\text{if} \quad \lambda + \rho z \geq 0;\\
        -\frac{\lambda^2}{2\rho} \quad &\text{otherwise}.
    \end{cases}
\end{equation}
\begin{equation}
    P_2(z,\rho,\lambda) = \begin{cases}
        \lambda z + \lambda \rho z^2+ \frac{1}{6}\rho^2z^3 &\text{if}\quad z\geq 0\\
        \frac{\lambda z}{1 - \rho z} &\text{if}\quad z \leq 0
    \end{cases}
\end{equation}
\begin{equation}
    P_3(z,\rho,\lambda) = \begin{cases}
        \lambda z + \lambda \rho z^2 &\text{if}\quad z\geq 0\\
        \frac{\lambda z}{1 - \rho z} &\text{if}\quad z \leq 0
    \end{cases}
\end{equation}

\section{Dataset description with implementation details}
\label{sup:dataset}

    \noindent \textbf{Tiny-ImageNet} \cite{deng2009imagenet} is a standard benchmark for image classification and commonly used in the calibration literature \cite{mukhoti2020calibrating, liu2022mbls}.
    It includes $64\times64$ dimensional images across $200$ classes, with $500$ images per class in the train set and $50$ per class in the validation set.
    Following \cite{mukhoti2020calibrating}, we split out a validation set by randomly choose $50$ samples per class from the train set, while the original validation set is used as the test set.
    We train ResNet-50 \cite{he2016deep} model by SGD optimizer with a batch size of $128$, and the number of epochs is set to 100 .
    A multi-step learning rate scheduling strategy is used, \ie learning rate 0.1 for the first 40 epochs, 0.01 for the next 20 epochs and 0.001 for the rest.
 
     \noindent \textbf{ImageNet} \cite{deng2009imagenet} is a large-scaled image classification benchmark. We use the version of ILSVRC-2012 (or ImageNet-1K) in our experiments (referred as ImageNet in this paper). It consists of 1K object classes with 1.2M images for training and $5$K for validation.
    The average resolution of an image is $469\times387$.
    We follow \cite{minderer2021revisiting} for evaluating calibration performance on ImageNet, \ie reserving $20\%$ for validation and the remaining $80\%$ for testing.
    Besides ResetNet-50 \cite{he2016deep}, we also train state-of-the-art transformer based network, \ie SwinV2-T \cite{liu2021swinv2}, on this dataset.
    AdamW \cite{loshchilov2018adamw} optimizer is applied, and a cosine learning rate scheduler \cite{loshchilov2017sgdr} with an initial learning rate of 0.001 is used.
    The number of training epochs is set to $200$ and $300$ for ResNet-50 and SwinV2-T respectively.
    The input size is $224\times224$ for ResNet-50 and $256\times256$ for SwinV2-T, while the batch size is $1024$ for training both networks.
    Regular data augmentation techniques like random resized crop, random horizontal flips, random color jitter, and random pixel erasing are applied on the training samples.
    
     \noindent \textbf{ImageNet-LT} \cite{Liu2019imagenetlt} is truncated from ImageNet by sampling a subset so that the labels of the training set follow a long-tailed distribution.
    Overall, it has $115.8$K images belonging to $1$K classes, and the number of samples per class ranges from $5$ to $1280$. Both the validation and test sets are balanced, where the validation set includes $20$ images per class and the original validation set in ImageNet is employed as the test set.
    Regarding the networks and training details, we use the same settings as those on ImageNet.

\begin{table*}[htb]
    \centering
    \begin{tabular}{l*{11}{c}}
        & \multicolumn{2}{c}{TinyImageNet} & \multicolumn{2}{c}{ImageNet} & \multicolumn{2}{c}{ImageNet-LT} & \multicolumn{2}{c}{20 News} \\
        \cmidrule(lr){2-3}\cmidrule(lr){4-5}\cmidrule(lr){6-7}\cmidrule(lr){8-9}
        Method & Pre-TS & Post-TS & Pre-TS & Post-TS & Pre-TS & Post-TS & Pre-TS & Post-TS \\
        \midrule
        CE                & $3.73$ & $1.86_{\,1.1}$ & $9.19$ & $3.88_{\,1.6}$ & $28.12$ & $3.72_{\,1.7}$ & $22.75$ & $3.01_{\,3.1}$ \\
        LS                & $3.17$ & $1.79_{\,0.9}$ & $2.57$ & $2.57_{\,1.0}$ & $10.46$ & $3.32_{\,1.3}$ & $ 8.07$ & $3.69_{\,1.2}$ \\
        FL                & $2.96$ & $1.74_{\,0.9}$ & $1.60$ & $1.60_{\,1.0}$ & $18.37$ & $2.52_{\,1.5}$ & $10.80$ & $3.33_{\,1.4}$ \\
        FLSD              & $2.91$ & $1.74_{\,0.9}$ & $2.08$ & $2.08_{\,1.0}$ & $17.77$ & $3.40_{\,1.4}$ & $10.87$ & $4.10_{\,1.4}$ \\
        CPC               & $4.88$ & $2.66_{\,1.5}$ & $3.66$ & $2.00_{\,1.1}$ & $16.00$ & $3.22_{\,1.2}$ & $ 9.46$ & $4.35_{\,1.4}$ \\
        MbLS              & $1.64$ & $1.64_{\,1.0}$ & $4.44$ & $2.07_{\,1.1}$ & $ 6.16$ & $2.60_{\,1.1}$ & $ 5.40$ & $2.09_{\,1.1}$ \\
        \midrule
        CALS-HR           & $2.50$ & $1.82_{\,0.9}$ & $5.63$ & $1.68_{\,1.4}$ & $ 2.83$ & $2.83_{\,1.0}$ & $ 6.99$ & $3.14_{\,1.1}$ \\
        \textbf{CALS-ALM} & $\mathbf{1.54}$ & $\mathbf{1.54}_{\,1.0}$ & $\mathbf{1.46}$ & $\mathbf{1.28}_{\,1.1}$ & $\mathbf{2.15}$ & $\mathbf{1.81}_{\,0.9}$ & $\mathbf{2.04}$ & $\mathbf{1.86}_{\,1.1}$  \\
        \bottomrule
    \end{tabular}
    \caption{Calibration performance (ECE in \%) when adding post temperature scaling (best T value for each method in subscript). The architecture is fixed to ResNet-50 for the vision datasets and GPCN for 20 News dataset.}
    \label{tab:temp}
\end{table*}

    \noindent \textbf{PASCAL VOC2012} \cite{VOC2015} is a natural semantic segmentation benchmark including $20$ foreground object classes and an additional background class.
    As the original test set is not publicly released and it is unable to evaluate the calibration performance via the official evaluation server, we split out a validation set by randomly selecting $20\%$ images from the training set and treat the original validation set as our test set.
    Overall, the training/validation/test split contains $1171/293/1449$ images.
    For segmentation model training, we employ DeepLabV3 \cite{chen2017rethinking} implemented by the popular public library\footnote{\url{https://github.com/qubvel/segmentation_models.pytorch}}, where we use ResNet-34 as encoder initialized with pre-trained weights on ImageNet, and the decoder is trained from scratch.
    The batch size is set to 8 and AdamW optimizer is used with an initial learning rate of 0.001 alongside a cosine learning rate scheduler.
    Finally, the maximum training epoch is set to 100.
    
     \noindent \textbf{20 Newsgroups} \cite{lang1995newsweeder}. To evaluate the generalization of the proposed method, we include a non-vision dataset, \ie 20 Newsgroups, which is a text classification benchmark and also used in previous calibration papers \cite{mukhoti2020calibrating, liu2022mbls}.
    It contains $20K$ news articles from $20$ different groups according to the content, \eg rec.motorcycles, rec.autos, sci.space, etc.
    We use the standard data split setting : $15,098$ documents for training, $900$ for validation and $3,999$ for testing.
    The Glove word embedding \cite{pennington2014glove} is used to encode the text and then a Global Pooling Convolutional Network (GPCN) \cite{lin2013network} is trained.
    During training, we use Adam optimizer with an initial learning rate of 0.001.
    We train the model for 100 epochs, where the learning rate is decayed by a factor of 0.1 after the first 50 epochs.

\section{Additional results}
\label{sup:add}

\autoref{tab:temp} reports the results of post-training temperature scaling (post-TS) on the outputs of the trained models \cite{guo2017calibration}.
Since this post-process technique is orthogonal to training based methods, we also present the results of applying it to our method, as well as the related works.
We can see that our method without temperature scaling (pre-TS) outperforms previous methods, even post-TS, across all the benchmarks.
Additionally, the ECE of our model is further reduced with post-TS in some cases, for instance on ImageNet (${1.46\% \to 1.28\%}$) and ImageNet-LT (${2.04\% \to 1.81\%}$).

\begin{table}
    \centering
    \begin{tabular}{l*{2}{c}}
        Method & ImageNet & ImageNet-LT \\
        \midrule
        CE                & $0.036$ & $0.090$ \\
        LS                & $0.029$ & $0.072$ \\
        FL                & $0.030$ & $0.087$ \\
        FLSD              & $0.029$ & $0.087$ \\
        CPC               & $0.049$ & $0.078$ \\
        MbLS              & $0.030$ & $0.072$ \\
        \midrule
        CALS-HR           & $0.029$ & $0.071$ \\
        \textbf{CALS-ALM} & $\mathbf{0.027}$ & $\mathbf{0.069}$   \\
        \bottomrule
    \end{tabular}
    \caption{Class-wise Calibration Error (CWCE in \%) computed for different approaches on ImageNet and ImageNet-LT. The architecture is fixed to ResNet-5. Best method is highlighted in bold.}
    \label{tab:cwce}
\end{table}

\autoref{tab:cwce} reports the performance on the two natural image datasets, \ie ImageNet and ImageNet-LT, in terms of Class-wise Calibration Errors (CWCE) \cite{MaierHein2022MetricsRP}, which is a class-wise extension of ECE. Our method consistently achieve the best scores, with relative improvements of $25.0\%$ on ImageNet and $23.3\%$ on ImageNet-LT.

\begin{table}[b]
  \centering
  \scriptsize 
  \begin{tabular}{@{}l*{5}{c}@{}}
    & CE & LS & FL & MbLS & \textbf{Ours} \\
    \midrule
    ImageNet $\rightarrow$ ImageNet-C & 26.25 & 24.00 & 23.73 & 26.55 & \textbf{22.52} \\
    ImageNet-LT $\rightarrow$ ImageNet-C & 19.99 & 27.51 & 15.80 & 15.40 & \textbf{12.91} \\
    \bottomrule
  \end{tabular}
  \vspace{-2mm}
  \caption{ECE (\%) on the out-of-distribution dataset, \ie ImageNet-C (Gaussian noise corruption with severity level 5), for models trained on in-distribution datasets, \ie ImageNet and ImageNetLT.
  }
  \label{table:ood}
  \vspace{-1mm}
\end{table}

In \autoref{table:ood}, we present results on the out-of-distribution (OOD) scenario \cite{minderer2021revisiting}.
It is shown nn both settings, our method achieves the lowest ECE on the target domain.
These results confirm the effectiveness of our method in the OOD scenario.

\section{Visualization of learned classwise multipliers}
\label{sup:vis-class-lambd}

\begin{figure*}
    \centering
    \includegraphics{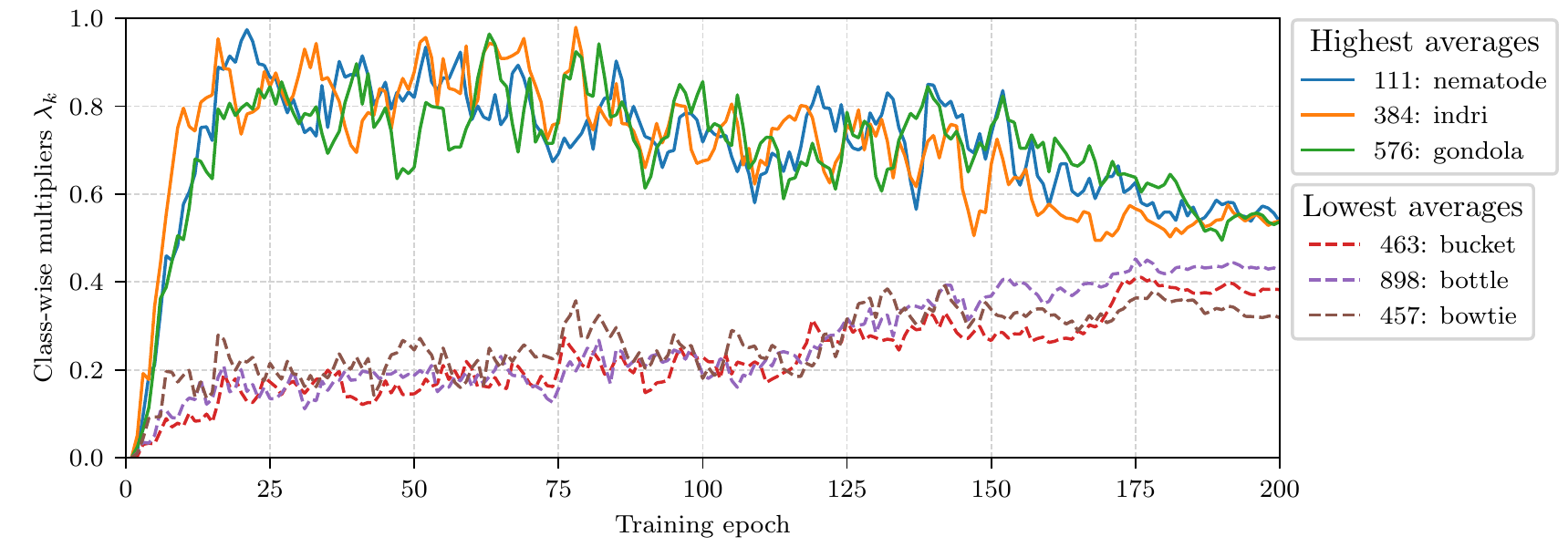}
    \caption{Visualization of learned multipliers $\lambda_k$ during the training of the ResNet-50 model on ImageNet. We show classes with the highest average (\textit{Solid lines}) and the lowest average (\textit{dashed lines}).}
    \label{fig:class_lambd}
\end{figure*}

\autoref{fig:class_lambd} shows the evolution of learned multipliers $\lambda_k$ on ImageNet for the three classes with the highest average and the three classes with the lowest average.
This highlights the advantages of our method: 1) assigning distinct penalty weights for different classes; 2) adaptively updating the weight for each class throughout the training process.

\section{Reliability diagram}
\label{sup:reliability}

\begin{figure*}
    \centering
    \includegraphics[width=1\linewidth]{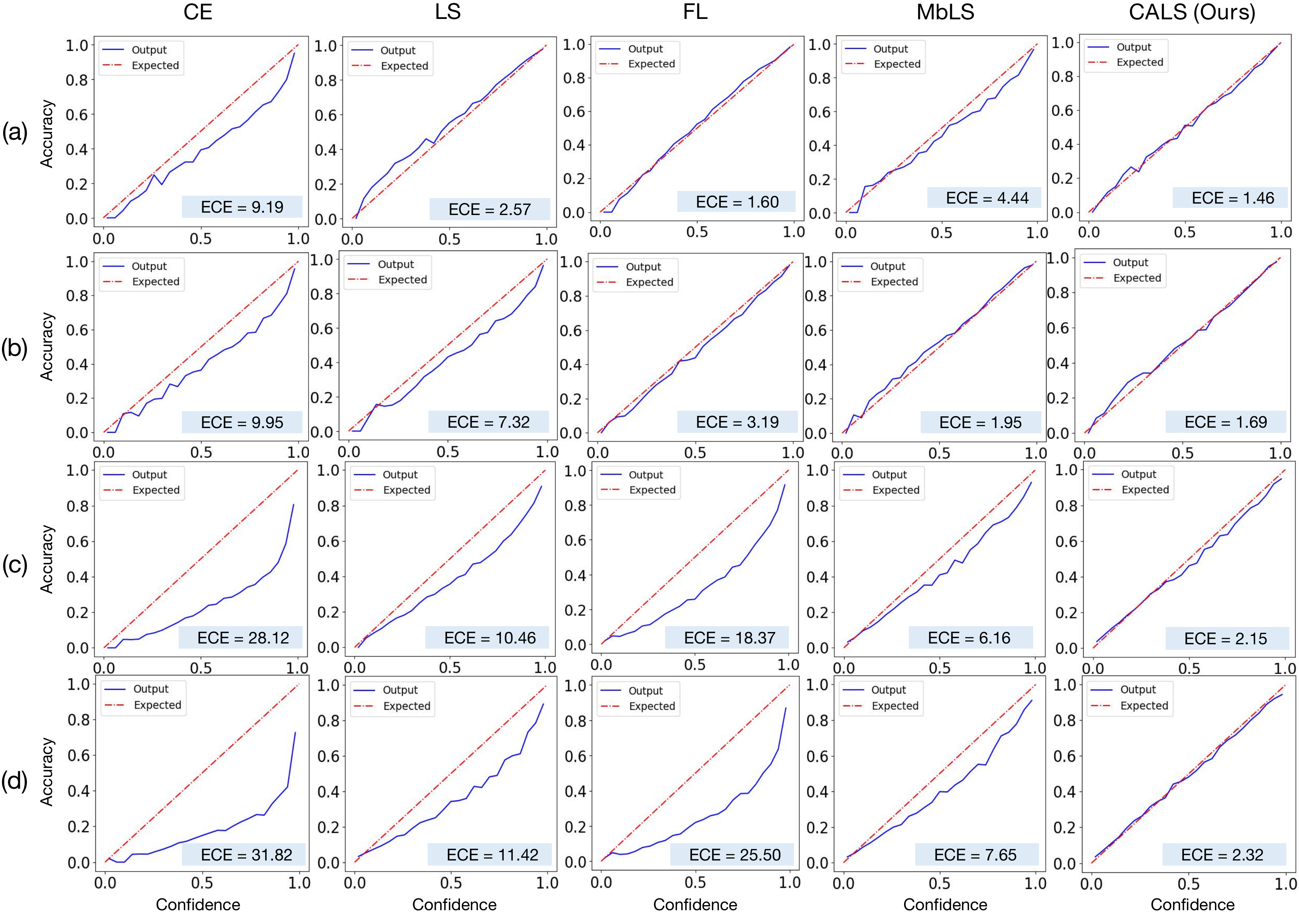}
    \caption{\textbf{Calibration visualizations: (a) ImageNet (ResNet-50), (b) ImageNet (SwinV2-T), (c) ImageNet-LT (ResNet-50), and (d) ImageNet-LT (SwinV2-T) .} We present the reliability diagrams of our method (CALS), compared with those of baselines and closely related works. The number of bins to plot reliability diagrams is set to 25.}
    \label{fig:reliable}
\end{figure*}

\autoref{fig:reliable} presents the reliability diagrams for different models trained on ImageNet and ImageNet-LT, which is a standard way of visualizing calibration performance.
The curve of a perfectly calibrated model in the reliability diagram should match the dashed red line, where the prediction confidence perfectly reflects the accuracy of the model. 
It is shown that the models trained with CE (\textit{left-most plots}) are over-confident, with accuracy mostly lower than confidence.
Our method, CALS, is the most effective one to pull the curves closer to the expected lines, showing nearly perfect calibration performances.
In particular, the improvement on ImageNet-LT is substantial compared to the other methods like LS and FL, which further demonstrates that the proposed class adaptive learning method could address the class imbalance issue in the long-tailed dataset.
On ImageNet, LS and FL also present strong calibration performance, but decrease the final accuracy as shown in Table 1 of the main text.
Overall, our method achieves the best compromise between calibration and accuracy.
It is noted that the observation from \autoref{fig:reliable} is supported by the quantitative scores reported in Table 1 of the main text.

\section{Hyper-parameter setting}
\label{sup:hyper}

\begin{table}
    \centering
    \begin{tabular}{@{}lcc@{}}
        Hyper-parameter & Value \\
        \midrule
        Margin $m$ (all vision tasks) & 10  \\
        Margin $m$ (text classification) & 6  \\
        Initial multiplier $\vlambda^{(0)}$ & $10^{-6}\cdot\1_K$ \\
        Initial Penalty parameter $\vrho^{(0)}$ & $\1_K$ \\
        Penalty increasing factor $\gamma$ & 1.2 \\
        Constraint improvement factor $\tau$ & 0.9 \\
        Period of penalty parameter update & 10 \\ 
        \bottomrule
    \end{tabular}
    \caption{Hyper-parameters for our method, \ie CALS-ALM.}
    \label{tab:hyper}
\end{table}

\autoref{tab:hyper} gives details of the hyper-parameter settings in our method, \ie CALS-ALM.
Note, the margin values are set by following \cite{liu2022mbls}, \ie 10 for all the vision tasks including classification and segmentation, and $6$ for the text classification on 20 Newsgroups.

Regarding the other methods in Table 1 of the main text, we set their hyper-parameters by following previous works, except that the values for MMCE\cite{kumar2018trainable} and CPC\cite{Cheng22CPC} are empirically set according to our implementation.
Detailed hyper-parameter settings for each method are as follows:
\begin{itemize}
\item MMCE \cite{kumar2018trainable}: balancing weight $\lambda=0.1$.
\item ECP \cite{pereyra2017regularizing}: balancing weight $\lambda=0.1$.
\item LS \cite{szegedy2016rethinking}: smoothing factor $\alpha=0.05$.
\item FL \cite{mukhoti2020calibrating}: scaling factor $\gamma=3$
\item FLSD \cite{mukhoti2020calibrating}: scaling factor $\gamma$ is set to $5$ for $s_k \in [0, 0.2)$ and $3$ for $s_k \in [0.2, 1)$, where $k$ is the right class for the sample.
\item CPC \cite{Cheng22CPC}: balancing weights for the binary discrimination penalty and binary exclusion penalty are set to $10$ and $1$ respectively. It is noted that we re-implement CPC since the official code is not publicly available.
\item MbLS \cite{liu2022mbls}: balancing weight $\lambda=0.1$, margin $m=10$ for all the vision tasks, and $m=6$ for the text classification task.
\end{itemize}

\end{document}